\documentclass{article}
\usepackage[english]{babel}
\usepackage[utf8]{inputenc}
\usepackage{listings}
\usepackage{johd}

\title{GNS: A generalizable Graph Neural Network-based simulator for particulate and fluid modeling}

\author{Krishna Kumar$^{a}$$^{*}$ and Joseph Vantassel$^{b}$. \\
        \small $^{a}$Department of Civil, Architectural and Environmental Engineering, University of Texas at Austin, TX, USA \\
        \small $^{b}$Department of Civil and Environmental Engineering, Virginia Tech, Virginia, USA\\\\
        \small $^{*}$Corresponding author: Krishna Kumar; \tt{krishnak@utexas.edu}
}

\date{} 

\begin{document}

\maketitle

\begin{abstract} 
\noindent We develop a PyTorch-based Graph Network Simulator (GNS) that learns physics and predicts the flow behavior of particulate and fluid systems.  GNS discretizes the domain with nodes representing a collection of material points and the links connecting the nodes representing the local interaction between particles or clusters of particles.  The GNS learns the interaction laws through message passing on the graph.  GNS has three components: (a) Encoder, which embeds particle information to a latent graph, the edges are learned functions; (b) Processor, which allows data propagation and computes the nodal interactions across steps; and (c) Decoder, which extracts the relevant dynamics (e.g., particle acceleration) from the graph.  We introduce physics-inspired simple inductive biases, such as an inertial frame that allows learning algorithms to prioritize one solution (constant gravitational acceleration) over another, reducing learning time.  The GNS implementation uses semi-implicit Euler integration to update the next state based on the predicted accelerations.  GNS trained on trajectory data is generalizable to predict particle kinematics in complex boundary conditions not seen during training.  The trained model accurately predicts within a 5\% error of its associated material point method (MPM) simulation.  The predictions are 5,000x faster than traditional MPM simulations (2.5 hours for MPM simulations versus 20 s for GNS simulation of granular flow).  GNS surrogates are popular for solving optimization, control, critical-region prediction for in situ viz, and inverse-type problems. The GNS code is available under the open-source MIT license at https://github.com/geoelements/gns.
\end{abstract}

\noindent\\

\section{Summary}
\begin{figure}
    \centering
    \includegraphics[width=\textwidth]{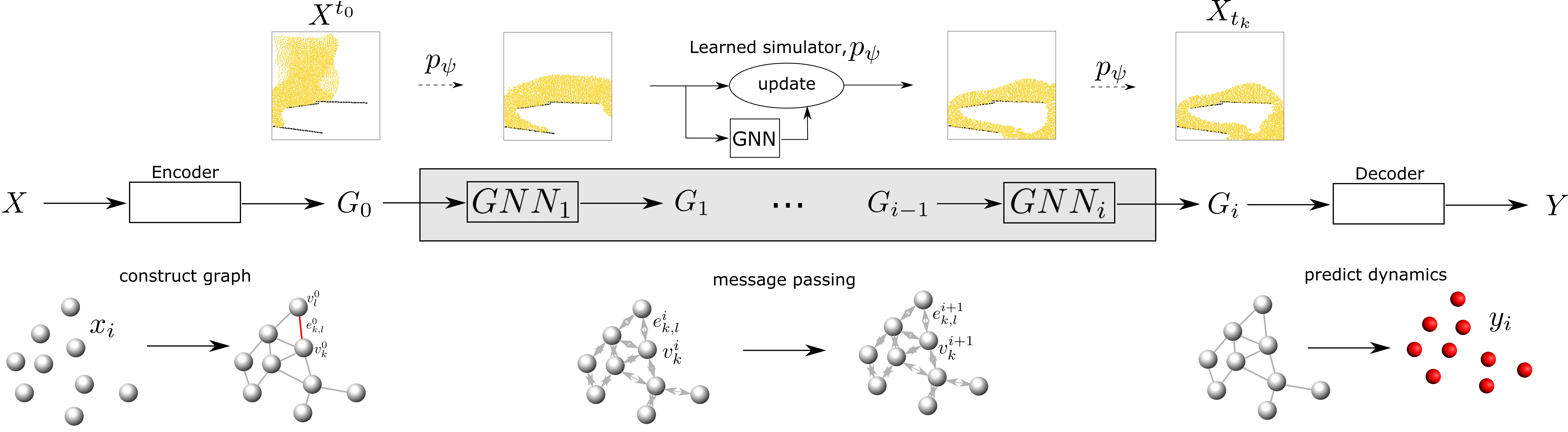}
    \caption{An overview of the graph network simulator (GNS).}
    \label{fig:gns}
\end{figure}

\noindent Graph Network-based Simulator (GNS) is a framework for developing generalizable, efficient, and accurate machine learning (ML)-based surrogate models for particulate and fluid systems using Graph Neural Networks (GNNs). GNNs are the state-of-the-art geometric deep learning (GDL) that operates on graphs to represent rich relational information~\cite{scarselli2008graph}, which maps an input graph to an output graph with the same structure but potentially different node, edge, and global
feature attributes. The graph network in GNS spans the physical domain with nodes representing an individual or a collection of particles, and the edges connecting the vertices representing the local interaction between particles or clusters of particles.  The GNS computes the system dynamics via learned message passing.  \Cref{fig:gns} shows an overview of how GNS learns to simulate n-body dynamics.  The GNS has three components: (a) Encoder, which embeds particle information to a
latent graph, the edges represent learned functions; (b) Processor, which allows data propagation and computes the nodal interactions across steps; and (c) Decoder, which extracts the relevant dynamics (e.g., particle acceleration) from the graph.  The GNS learns the dynamics, such as momentum and energy exchange, through a form of messages passing~\citep{gilmer2017neural}, where latent information propagates between nodes via the graph edges. The GNS edge messages (\(e^\prime_k \leftarrow \phi^e(e_k, v_{r_k}, v_{s_k}, u)\)) are a learned linear combination of the interaction forces.  The edge messages are aggregated at every node exploiting the principle of superposition \(\bar{e_i^\prime} \leftarrow \sum_{r_k = i} e_i^\prime\). The node then encodes the connected edge features and its local features using a neural network: \(v_i^\prime \leftarrow \phi^v (\bar{e_i}, v_i, u)\). 

\begin{figure}
\centering
\includegraphics[width=.8\textwidth]{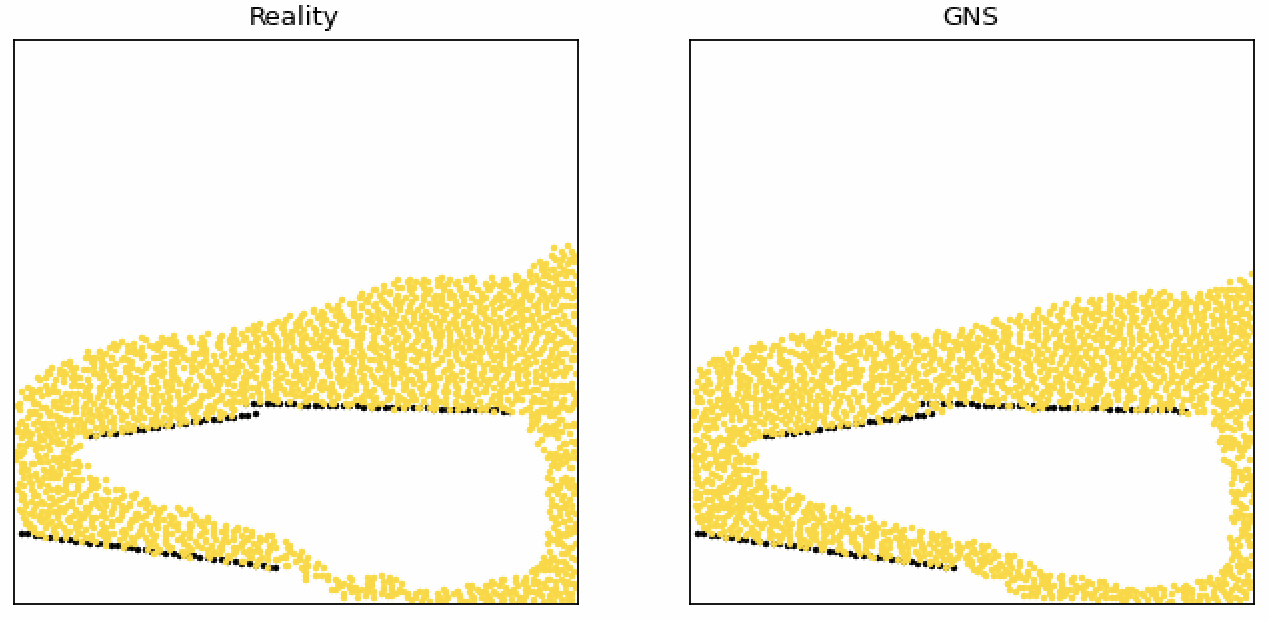}
\caption{GNS prediction of granular flow on ramps, compared against MPM simulation.\label{fig:gns-mpm}}
\end{figure}

The GNS implementation uses semi-implicit Euler integration to update the state of the particles based on the nodes predicted accelerations.  We introduce physics-inspired simple inductive biases, such as an inertial frame that allows learning algorithms to prioritize one solution over another (instead of learning to predict the inertial motion, the neural network learns to trivially predict a correction to the inertial trajectory, reducing learning time.  We developed an open-source, PyTorch-based GNS that predicts the dynamics of fluid and particulate systems~\citep{Kumar_Graph_Network_Simulator_2022}.  GNS trained on trajectory data is generalizable to predict particle kinematics in complex boundary conditions not seen during training.  \Cref{fig:gns-mpm} shows the GNS prediction of granular flow around complex obstacles trained on 20 million steps with 40 trajectories on NVIDIA A100 GPUs.  The trained model accurately predicts within 5\% error of its associated material point method (MPM) simulation.  The predictions are 5,000x faster than traditional MPM simulations (2.5 hours for MPM simulations versus 20 s for GNS simulation of granular flow) and are widely used for solving optimization, control and inverse-type problems.  In addition to surrogate modeling, GNS trained on flow problems is also used as an oracle to predict the dynamics of flows to identify critical regions of interest for in situ rendering and visualization~\citep{kumar2022insitu}.  The GNS code is distributed under the open-source MIT license and is available on~\url{https://github.com/geoelements/gns}.

\section{Statement of need}
Traditional numerical methods for solving differential equations are invaluable in scientific and engineering disciplines.  However, such simulators are computationally expensive and intractable for solving large-scale and complex inverse problems, multiphysics, and multi-scale mechanics.  Surrogate models trade off generality for accuracy in a narrow setting.  Recent growth in data availability has spurred data-driven machine learning (ML) models that train directly from observed data~\citep{prume2022model}.  ML models require significant training data to cover the large state space and complex dynamics.  Instead of ignoring the vast amount of structured prior knowledge (physics), we can exploit such knowledge to construct physics-informed ML algorithms with limited training data.  GNS uses static and inertial priors to learn the interactions between particles directly on graphs and can generalize with limited training data~\citep{wu2020comprehensive,velivckovic2017graph}.  Graph-based GNS offer powerful data representations of real-world applications, including particulate systems, material sciences, drug discovery, astrophysics, and engineering~\citep{sanchez2020learning,battaglia2018relational}.

\section{State of the art}

\citet{sanchez2020learning} developed a reference GNS implementation based on TensorFlow v1~\citep{tensorflow2015whitepaper}.  Although the reference implementation runs both on CPU and GPU, it doesn't achieve multi-GPU scaling.  Furthermore, the dependence on TensorFlow v1 limits its ability to leverage features such as eager execution in TF v2.  We develop a scalable and modular GNS using PyTorch using the Distributed Data Parallel model to run on multi-GPU systems.

\section{Key features}

The Graph Network Simulator (GNS) uses PyTorch and PyTorch Geometric for constructing graphs and learned message passing. GNS is highly-scalable to 100,000 vertices and more than a million edges. The PyTorch GNS supports the following features:

\begin{itemize}
    \item CPU and GPU training
    \item Parallel training on multi-GPUs
    \item Multi-material interactions
    \item Complex boundary conditions
    \item Checkpoint restart
    \item VTK results
    \item Animation postprocessing
\end{itemize}

\section{GNS training and prediction}

GNS models are trained on 1000s of particle trajectories from MPM (for sands) and Smooth Particle Hydrodynamics (for water) for 20 million steps. 

\subsection{Dataset format}

We use the numpy \verb|.npz| format for storing positional data for GNS training.  The \verb|.npz| format includes a list of tuples of arbitrary length where each tuple corresponds to a differenet training trajectory and is of the form \verb|(position, particle_type)|.  The data loader provides \verb|INPUT_SEQUENCE_LENGTH| positions, set equal to six by default, to provide the GNS with the last \verb|INPUT_SEQUENCE_LENGTH| minus one positions as input to predict the position at the next time step.  The `position' is a 3-D tensor of shape \verb|(n_time_steps, n_particles, n_dimensions)| and \verb|particle_type| is a 1-D tensor of shape \verb|(n_particles)|.

\noindent The dataset contains:

\begin{itemize}
    \item Metadata file with dataset information ``\textit{(sequence length, dimensionality, box bounds, default connectivity radius, statistics for normalization, ...)}":
    \begin{lstlisting}
    {
      "bounds": [[0.1, 0.9], [0.1, 0.9]], 
      "sequence_length": 320, 
      "default_connectivity_radius": 0.015, 
      "dim": 2, 
      "dt": 0.0025, 
      "vel_mean": [5.123277536458455e-06, -0.0009965205918140803], 
      "vel_std": [0.0021978993231675805, 0.0026653552458701774], 
      "acc_mean": [5.237611158734309e-07, 2.3633027988858656e-07], 
      "acc_std": [0.0002582944917306106, 0.00029554531667679154]
    }
    \end{lstlisting}
    
    \item  npz containing data for all trajectories `\textit{(particle types, positions, global context, ...)}'.
\end{itemize} 

Training datasets for Sand, SandRamps, and WaterDropSample are available on DesignSafe Data Depot \url{https://www.designsafe-ci.org/data/browser/public/designsafe.storage.published/PRJ-3702}~\citep{vantassel2022gnsdata}.

\section{Parallelization and scaling}

\begin{figure}
\centering
    \includegraphics[width=.8\textwidth]{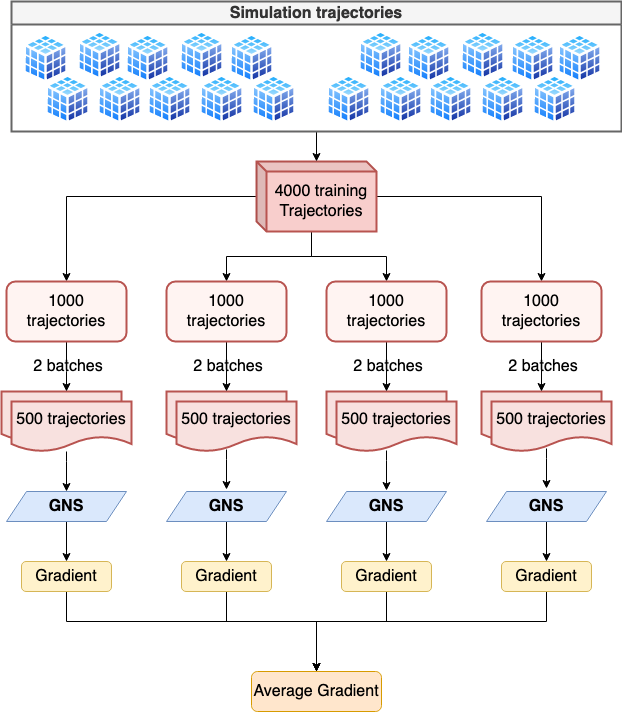}
    \caption{Distributed data parallelization in GNS.}
    \label{fig:gns-ddp}
\end{figure}

The GNS is parallelized to run across multiple GPUs using the PyTorch Distributed Data Parallel (DDP) model.  The DDP model spawns as many GNS models as the number of GPUs, distributing the dataset across all GPU nodes.  Consider, our training dataset with 20 simulations, each with 206 time steps of positional data $x_i$, which yields $(206 - 6) \times 20 = 4000$ training trajectories.  We subtract six position from the GNS training dataset as we utilize five previous velocities, computed from six positions, to predict the next position.  The 4000 training tajectories are subsequently distributed equally to the four GPUs (1000 training trajectories/GPU).  Assuming a batch size of 2, each GPU handles 500 trajectories in a batch.  The loss from the training trajectories are computed as

$$f(\theta) = \frac{1}{n}\sum_{i=1}^n (GNS_\theta(x_t^i) - a_t^i)\,,$$

where $n$ is the number of particles (nodes) and $\theta$ is the learnable parameter in the GNS. In DDP, the gradient $\nabla (f(\theta))$ is computed as the average gradient across all GPUs as shown in~\cref{fig:gns-ddp}.

\begin{figure}
    \centering
    \includegraphics[width=.8\textwidth]{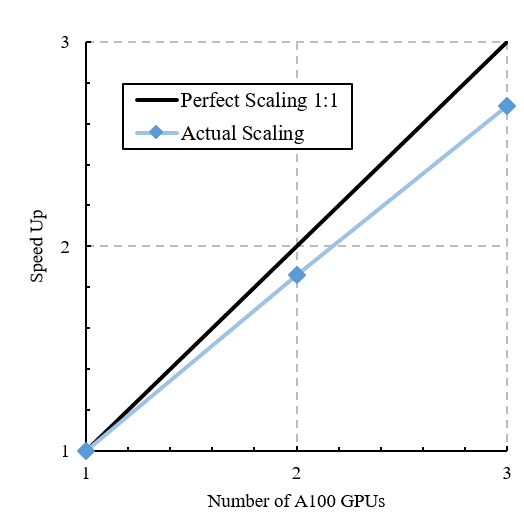}
    \caption{GNS strong-scaling on up to three NVIDIA A100 GPUs.}
    \label{fig:gns-scaling}
\end{figure}

A test of the GNS's scalability was performed on a node of Lonestar 6 at the Texas Advanced Computing Center equipped with three NVIDIA A100 GPUs.  Performance wave evaluated using the WaterDropSample training dataset for 6000 training steps.  Tests were performed using the recommended `nccl` DDP backend.  Results of the strong-scaling test, see~\cref{fig:gns-scaling}, show strong scaling performance.

\section*{Funding Statement}
We acknowledge the support of National Science Foundation NSF OAC: 2103937.

\bibliographystyle{apalike}
\bibliography{references}

\begin{thebibliography}{}

\bibitem[Abadi et~al., 2015]{tensorflow2015whitepaper}
Abadi, M., Agarwal, A., Barham, P., Brevdo, E., Chen, Z., Citro, C., Corrado,
  G.~S., Davis, A., Dean, J., Devin, M., Ghemawat, S., Goodfellow, I., Harp,
  A., Irving, G., Isard, M., Jia, Y., Jozefowicz, R., Kaiser, L., Kudlur, M.,
  Levenberg, J., Man\'{e}, D., Monga, R., Moore, S., Murray, D., Olah, C.,
  Schuster, M., Shlens, J., Steiner, B., Sutskever, I., Talwar, K., Tucker, P.,
  Vanhoucke, V., Vasudevan, V., Vi\'{e}gas, F., Vinyals, O., Warden, P.,
  Wattenberg, M., Wicke, M., Yu, Y., and Zheng, X. (2015).
\newblock {TensorFlow}: Large-scale machine learning on heterogeneous systems.
\newblock Software available from tensorflow.org.

\bibitem[Battaglia et~al., 2018]{battaglia2018relational}
Battaglia, P.~W., Hamrick, J.~B., Bapst, V., Sanchez-Gonzalez, A., Zambaldi,
  V., Malinowski, M., Tacchetti, A., Raposo, D., Santoro, A., Faulkner, R.,
  et~al. (2018).
\newblock Relational inductive biases, deep learning, and graph networks.
\newblock {\em arXiv preprint arXiv:1806.01261}.

\bibitem[Gilmer et~al., 2017]{gilmer2017neural}
Gilmer, J., Schoenholz, S.~S., Riley, P.~F., Vinyals, O., and Dahl, G.~E.
  (2017).
\newblock {Neural Message Passing for Quantum Chemistry}.
\newblock In {\em {International Conference on Machine Learning}}, pages
  1263--1272. PMLR.

\bibitem[Kumar et~al., 2022]{kumar2022insitu}
Kumar, K., Navratil, P., Solis, A., and Vantassel, J. (2022).
\newblock Minority report: A graph-network oracle for large-scale in situ
  visualization.
\newblock In {\em IEEE Large-Scale Data Analysis and Visualization}, Oklahoma,
  USA. IEEE.

\bibitem[Kumar and Vantassel, 2022]{Kumar_Graph_Network_Simulator_2022}
Kumar, K. and Vantassel, J. (2022).
\newblock {Graph Network Simulator: v1.0.1}.

\bibitem[Prume et~al., 2022]{prume2022model}
Prume, E., Reese, S., and Ortiz, M. (2022).
\newblock Model-free data-driven inference in computational mechanics.
\newblock {\em arXiv preprint arXiv:2207.06419}.

\bibitem[Sanchez-Gonzalez et~al., 2020]{sanchez2020learning}
Sanchez-Gonzalez, A., Godwin, J., Pfaff, T., Ying, R., Leskovec, J., and
  Battaglia, P. (2020).
\newblock Learning to simulate complex physics with graph networks.
\newblock In {\em International Conference on Machine Learning}, pages
  8459--8468. PMLR.

\bibitem[Scarselli et~al., 2008]{scarselli2008graph}
Scarselli, F., Gori, M., Tsoi, A.~C., Hagenbuchner, M., and Monfardini, G.
  (2008).
\newblock The graph neural network model.
\newblock {\em IEEE transactions on neural networks}, 20(1):61--80.

\bibitem[Vantassel and Kumar, 2022]{vantassel2022gnsdata}
Vantassel, J. and Kumar, K. (2022).
\newblock {Graph Network Simulator: v1.0.1}.

\bibitem[Veli{\v{c}}kovi{\'c} et~al., 2017]{velivckovic2017graph}
Veli{\v{c}}kovi{\'c}, P., Cucurull, G., Casanova, A., Romero, A., Lio, P., and
  Bengio, Y. (2017).
\newblock Graph attention networks.
\newblock {\em arXiv preprint arXiv:1710.10903}.

\bibitem[Wu et~al., 2020]{wu2020comprehensive}
Wu, Z., Pan, S., Chen, F., Long, G., Zhang, C., and Philip, S.~Y. (2020).
\newblock A comprehensive survey on graph neural networks.
\newblock {\em IEEE Transactions on Neural Networks and Learning Systems}.

\end{thebibliography}

\end{document}